\pdfoutput=1

\documentclass[11pt]{article}

\usepackage{ACL2023}

\usepackage{times}
\usepackage{latexsym}

\usepackage[T1]{fontenc}

\usepackage[utf8]{inputenc}

\usepackage{microtype}

\usepackage{inconsolata}

\usepackage{multirow}
\usepackage{graphicx}
\usepackage{ulem}
\usepackage{arydshln}
\newcommand{\naturaltwo}{\textsc{Super-NaturalInstru}}
\newcommand{\exactmatch}{\textsc{ExactMatch}}
\newcommand{\rougel}{\textsc{ROUGE-L}}
\newcommand{\longinstruc}{\textsc{Instru-w-Exam}}
\newcommand{\shortinstruc}{\textsc{Instru-wo-Exam}}
\newcommand{\diverseprompt}{\textsc{DiversePrompt}}
\definecolor{chocolate}{rgb}{0.82, 0.41, 0.12}

\usepackage{tabu}
\usepackage{xcolor}

\title{Robustness of Learning from Task Instructions}

\author{
   Jiasheng Gu$^{\dagger}$, Hongyu Zhao$^\S$, Hanzi Xu$^\star$, Liangyu Nie$^\diamond$, Hongyuan Mei$^\P$,  Wenpeng Yin$^\ddag$
   \\
   $^\dagger$University of Southern California; $^\S$University of Chicago; $^\star$Temple University\\ 
   $^\diamond$University of Birmingham; $^\P$Toyota Technological Institute at Chicago; $^\ddag$Penn State University\\
   \texttt{gujiashe@usc.edu}; $\;$\texttt{hongyuz@uchicago.edu}; $\;$\texttt{abdielnie@gmail.com}\\\texttt{hanzi.xu@temple.edu}; $\;$\texttt{hongyuan@ttic.edu}; \texttt{wenpeng@psu.edu}\\
 }

\begin{document}
\maketitle
\begin{abstract}
Traditional supervised learning mostly works on individual tasks and requires training on a large set of task-specific examples. This paradigm seriously hinders the development of task generalization since preparing a task-specific example set is costly. To build a system that can quickly and easily generalize to new tasks, textual instructions have been adopted as an emerging trend of supervision recently. These instructions give the model the definition of the task and allow the model to output the appropriate answer based on the instructions and inputs. However, task instructions are often expressed in different forms, which can be interpreted from two threads \cite{DBLP10475}: firstly, some instructions are short sentences and are pretrained language model (PLM) oriented, such as prompts, while other instructions are paragraphs and are human-oriented, such as those annotation guidelines in Amazon MTurk; secondly, different end-users very likely explain the same task with instructions of different textual expressions. A robust system for task generalization should be able to handle any new tasks regardless of the variability of instructions.  

However, the system robustness in dealing with a wide range of instructions is still unexplored. This work investigates the system robustness when the instructions of new tasks are (i) manipulated, (ii) paraphrased, or (iii) from different levels of conciseness. To our knowledge, this is the first work that systematically studies how robust a PLM is when it is supervised by instructions with different factors of variability.\footnote{All code and data: \url{https://github.com/jiashenggu/Tk-Instruct/tree/main}}
\end{abstract}

\section{Introduction}

One goal of AI is to achieve cross-task generalization, i.e., the model can handle  new tasks after being trained on existing tasks. To solve target tasks, the system needs task-specific supervision. In the conventional machine learning paradigm, task-specific supervision means a large set of labeled examples particularly for this task. Despite the progress of conventional machine learning, collecting task-specific examples of large volumes is costly and even infeasible if we want the system to cope with new tasks immediately. Therefore, another form of task-specific supervision---task instructions---is attracting increasing attention in the community. Task instructions, commonly a piece of text, describe the task semantics and tell the model how to approach the expected outputs for the inputs. A couple of recent studies \cite{mishra-etal-2022-cross,yin-etal-2022-contintin,DBLP07705,T0} have shown promising progress in cross-task generalization with only instructions as supervision.

In the real world, task instructions are often presented in different forms, which can be interpreted from two dimensions: on the one hand, as \newcite{DBLP10475} defined,  \textit{instructions have different degrees of conciseness and were created for different audiences}---some instructions are short sentences and are \textbf{\texttt{PLM-oriented}}, such as prompts like ``translate from English to French'', while other instructions are paragraphs and are \textbf{\texttt{human-oriented}}, e.g., those annotation guidelines in Amazon MTurk; on the other hand, the same task is likely to be explained by different end-users  with instructions of  varying textual expressions. For example, ``translate from English to French'' can also be expressed as ``given an English sentence, translate it into French''. Therefore, a robust system for task generalization is expected to handle any new tasks regardless of the variability of instructions.  

However, despite the limited  earlier work studying the PLM robustness in terms of prompts \cite{Webson2021-zn,DBLPKhashabiBCH22}, it is still unclear how robust a pretrained instruction learning system can handle  broader types of instructions, including \texttt{human-oriented} as well as \texttt{PLM-oriented} ones, and their cross-type generalization. 
This work focuses on \texttt{human-oriented} instructions and explores the system robustness when the instructions of new tasks are (i) manipulated, e.g., words removal, words replacement, sentence shuffle, etc., (ii) paraphrased with dramatic surface change, or (iii) from different levels of conciseness, e.g., training tasks have  \texttt{human-oriented} instructions while testing tasks are equipped with \texttt{PLM-oriented} prompts. 

To get the answers for the above three threads of instruction variability, we finetune a T5 \cite{DBLPRaffelSRLNMZLL20} model on the $train$ of \naturaltwo~\cite{DBLP07705}, a dataset with the largest number of NLP tasks and each task is described by an Amazon-Mturk-style instruction, and test it on (i) the $test$ of \naturaltwo,~where we modify the original instructions with word-level, sentence-level and instruction-level perturbations, (ii) a newly crowdsourced dataset, \texttt{Para-Instructions},  with 23 unseen tasks and each has 4 distinct \texttt{human-oriented} instructions in the same format as the $test$ of \naturaltwo, and (iii) a new dataset \diverseprompt~\cite{Honovich_undated-gv}, in which each task has prompt-style instructions.

To our knowledge, this is the first work that systematically studies how robust a PLM is when it is supervised by instructions with different factors of variability. Our experiments show that:
\begin{itemize}
    \item The presence of demonstrations minimizes the impact of instruction perturbations and paraphrasing, as they typically have an unpredictable effect on performance. Demonstrations play a crucial role in ensuring system robustness, even when the instruction provided is a random text fragment, i.e.,  ``random instruction + positive examples'' mostly outperforms ``correct instruction only'';
    \item When demonstrations are absent, applying a model that learns from training tasks with  \texttt{human-oriented} instructions to unseen tasks  with  \texttt{PLM-oriented} prompts works poorly;

    \item So far, we still lack a method to effectively learn instruction to solve tasks in the absence of demonstration.
\end{itemize}

\begin{table*}[t]
\centering

\begin{tabu}{cl|p{11cm}}
\hline
\multicolumn{2}{c|}{\multirow{1}{*}{Definition}} & In this task, you will be shown an incorrect English sentence. You need to generate a corrected form of the input sentence.  \\
\hline

\multirow{3}{*}{\rotatebox{90}{\begin{tabu}{c} Positive\end{tabu}}}
& \multirow{1}{*}{input} & The car's wheel are loose. \\\cdashline{2-3}

& \multirow{1}{*}{output} & The car's wheel is loose.\\\cdashline{2-3}

& \multirow{1}{*}{explanation} & The instance of are is replaced by the word is, because the wheel is a singular word. This makes the sentence grammatically correct.\\

\hline
\multirow{3}{*}{\rotatebox{90}{\begin{tabu}{c} Negative\end{tabu}}}
& \multirow{1}{*}{input} & This way is the way to go. \\\cdashline{2-3}

& \multirow{1}{*}{output} & This way may be the way to go. \\\cdashline{2-3}

& \multirow{1}{*}{explanation} & The example does not correct the misuse of the word way. Instead, it should shorten the sentence to: this is the way to go.\\
\hline
\multirow{2}{*}{\rotatebox{90}{\begin{tabu}{c} Instances\end{tabu}}}
& \multirow{1}{*}{input} & For example , nobody is going to ask his personal doctor which he sees when he has a flue if he can also do a heart surgery or transplant organs .  
\\\cdashline{2-3}

& \multirow{1}{*}{output} & For example , nobody is going to ask his personal doctor , which he sees when he has a flu , if he can also do a heart surgery or transplant organs .   \\
\hline
\end{tabu}

\caption{A task sample (task1557 jfleg grammar error correction) from \naturaltwo: each task has a definition, a couple of positive examples, a couple of negative examples, and a large set of instances.}
\label{tab:naturalv2}
\end{table*}
\begin{table}[t]
 \setlength{\belowcaptionskip}{-10pt}
 \setlength{\abovecaptionskip}{5pt}
\centering
\begin{tabu}{clc}
\hline
metric &\textbf{Task Category} & \textbf{Abbr.}\\
\hline
\multirow{6}{*}{\rotatebox{90}{\begin{tabular}{c} Exact Match\end{tabular}}} &Textual Entailment & TE \\
&Cause Effect Classification & CEC \\
&Coreference Resolution & CR\\
&Dialogue Act Recognition & DAR \\ 
&Answerability Classification  & AC \\
&Word Analogy & WA  \\ 
\hline
\multirow{6}{*}{\rotatebox{90}{\begin{tabular}{c} ROUGE-L\end{tabular}}} & Overlap Extraction & OE \\ 
&Keyword Tagging & KT \\ 
&Question Rewriting & QR \\ 
&Title Generation & TG \\ 
&Data to Text & DTT \\ 
&Grammar Error Correction & GEC \\ \hline
\end{tabu}

\caption{12 task categories in the $test$ of \naturaltwo~and their official evaluation metrics.}
\label{tab:abbreviation}
\end{table}

\section{Related work}
Our work is related to prompts (a special case of instruction learning),  cross-task generalization, and the study of robustness under adversarial attack.

\paragraph{Prompt as instruction.}
\citet{Brown2020-qp} attempted to unleash the power of PLMs, such as GPT-3 \cite{DBLPBrownMRSKDNSSAA20}, through prompts. Since they discovered prompts, many studies have followed that direction in various ways, with robustness being an aspect that cannot be ignored. \citet{Liu2021-bz} introduced the effectiveness of prompt learning and the wide application prospects. \newcite{rethinking} explored how the model learns prompts and what aspects of the prompt contribute to the final task performance. \citet{Webson2021-zn} tried to figure out whether prompt-based models truly understand the meaning of their prompts. \citet{Zhao2021-cj} investigated ways to improve the stability of prompt-based models. \citet{bach-etal-2022-promptsource} created a platform for the production, distribution, and use of natural language prompts. Despite the large number of papers on prompts, little attention has been paid to studying the robustness of systems under prompt learning. Our work has a broader vision that considers the robustness in terms of \texttt{PLM-oriented } prompts as well as more natural \texttt{human-oriented} instructions.

\paragraph{Instruction-driven task generalization.}
Recent work has worked on using the information in task instructions to allow PLMs to respond to inputs that match the task instructions after they have understood  task instructions. \citet{DBLP982} tested
GPT-2 \citep{radford2019language} to understand real-world MTurk instructions to annotate some popular datasets, and concluded that
GPT-2 works poorly. \citet{DBLP08773} collected more than 60 distinct NLP tasks with  MTurk instructions consisting of items like \texttt{title}, \texttt{definition}, \texttt{things to avoid}, etc., and claimed that  BART \citep{lewisbart} and GPT-3  benefit from instructions to generalize across tasks. \citet{DBLP07705} further extended this dataset to cover over 1.6k cross-lingual tasks, each with a piece of natural language instruction. More  work in this thread can be found in \cite{DBLP10475}, which  provides a comprehensive survey for various instruction learning. The robustness studied in this work is also based on instruction-driven task generalization; however, our emphasis lies on analyzing the variability in system performance as opposed to actively enhancing overall system performance.

\paragraph{Robustness under adversarial attack.} There is a lot of work on adversarial attacks in NLP systems, and adversarial attacks are often used as a way to enhance robustness \cite{Zhang2019-sb,DBLPGoelRVTBR21,DBLPWangWQPLCBC20,DBLPMorrisLYGJQ20,DBLP07320}. Adversarial perturbations are considered to be worst-case scenarios that do not often occur in the real world and represent a very specific kind of noise. To generate valid adversarial examples, most attack methods require access to the structure, internal weights, and hyperparameters of the NLP model. In addition, adversarial perturbations should not be perceived by humans. This is a serious challenge because even small changes in the text may be easily recognized by the user but ignored by the machine. In contrast, some minor changes may cause the machine to react drastically but are not important to humans. Our investigation encompasses not only deliberate perturbations, but also  more natural discrepancies that arise between the training instructions and test instructions.

\begin{table*}[!ht]
 \setlength{\belowcaptionskip}{-10pt}
 \setlength{\abovecaptionskip}{5pt}
\centering

\begin{tabu}{cl|p{11cm}}
\hline
\multicolumn{2}{c|}{\multirow{1}{*}{Original Instruction}} & In this task, you will be shown an incorrect English sentence. You need to generate a corrected form of the input sentence.  \\\hline

\multirow{5}{*}{\rotatebox{90}{\begin{tabular}{c} word-level\end{tabular}}}& \multirow{1}{*}{Delete words} & In this \xout{task}, you \xout{will} be \xout{shown} an incorrect English sentence. You need to generate a corrected \xout{form} of \xout{the} input sentence.  \\\cdashline{2-3}

& \multirow{1}{*}{Delete stopwords} & \xout{In} \xout{this} task, \xout{you will be} shown \xout{an} incorrect English sentence. \xout{You} need \xout{to} generate \xout{a} corrected form \xout{of the}  input sentence.  \\\cdashline{2-3}

& \multirow{1}{*}{Insert words} & In this \underline{,} task, \underline{,} you will be shown an incorrect \underline{(} English \underline{input} sentence. You need to generate a corrected form of the \underline{English} input sentence. \\\cdashline{2-3}

& \multirow{1}{*}{Replace words} & In this \underline{case}, you will be shown an incorrect English \underline{language}. You need to generate a corrected form of the input sentence.  \\\cdashline{2-3}

& \multirow{1}{*}{Shuffle words} & a You generate English incorrect be need task, In the will this you sentence. shown to sentence. input an form of corrected \\
\hline
\multirow{2}{*}{\rotatebox{90}{\begin{tabu}{c} sent-level\end{tabu}}}&\multirow{1}{*}{Repeat sentences} & In this task, you will be shown an incorrect English sentence. You need to generate a corrected form of the input sentence. \emph{You need to generate a corrected form of the input sentence.} \\\cdashline{2-3}

& \multirow{1}{*}{Shuffle sentences} & You need to generate a corrected form of the input sentence. In this task, you will be shown an incorrect English sentence.  \\\hline
\multicolumn{2}{c|}{\multirow{1}{*}{instruction-level shuffle}} & In this task, you are given a sentence and a question, you would be asked to create the answer which is contained in the sentence provided.\\
\hline
\end{tabu}

\caption{Examples of original instructions, word-level, sentence-level, and instruction-level perturbations.}
\label{tab:perturbation_examples}
\end{table*}

\begin{table*}[!ht]
 \setlength{\belowcaptionskip}{-10pt}
 \setlength{\abovecaptionskip}{5pt}
\centering

\begin{tabu}{l|l|lllllll}
\hline
{metric} & &        AC &       CEC &        CR &       DAR &        WA &        TE \\
\hline
\multirow{18}{*}{\rotatebox{90}{\begin{tabu}{c} \exactmatch\end{tabu}}} & \multirow{2}{*}{original instru.   }          &  49.8±0.6 &  50.0±0.8 &  36.1±0.5 &  43.2±0.8 &  19.7±0.7 &  40.1±0.8 \\
\rowfont{\color{gray}} & & 63.6±1.2 &  48.2±3.0 &  48.1±0.9 &  57.0±1.2 &  51.8±0.4 &  46.6±0.5 \\\cdashline{3-8}
 & \multirow{2}{*}{\textcolor{cyan}{delete stopwords} }    &  44.2±0.4 &  45.0±1.2 &  32.3±0.7 &  28.7±0.9 &   7.60±0.1 &  31.6±0.5 \\
 \rowfont{\color{gray}}& & 63.2±0.6 &  49.0±0.5 &  49.1±2.2 &  49.9±1.7 &  50.5±1.9 &  44.5±0.6 \\\cdashline{3-8}
 & \multirow{2}{*}{\textcolor{cyan}{delete [5,10] words } }    &  45.3±3.3 &  40.8±4.3 &  32.5±0.4 &  34.0±2.6 &  18.6±2.6 &  34.9±3.1 \\
 \rowfont{\color{gray}}& & 61.1±2.0 &  50.7±1.2 &  49.5±1.3 &  52.9±4.5 &  50.8±1.3 &  44.0±1.4 \\\cdashline{3-8}
 & \multirow{2}{*}{\textcolor{cyan}{insert [5,10] words}  }    &  44.5±1.9 &  40.3±4.2 &  34.0±2.0 &  38.9±1.2 &  18.5±1.2 &  35.1±0.4 \\
\rowfont{\color{gray}} & & 60.4±1.4 &  48.7±1.3 &  47.5±0.9 &  56.6±0.9 &  52.5±1.6 &  44.6±0.5 \\\cdashline{3-8}
 & \multirow{2}{*}{\textcolor{cyan}{replace [5,10] words } }   &  48.7±1.0 &  45.9±1.6 &  31.3±1.9 &  35.9±1.8 &  18.2±0.6 &  33.8±1.7 \\
\rowfont{\color{gray}} & & 60.3±0.7 &  49.2±1.5 &  47.7±1.0 &  50.0±0.9 &  51.6±1.9 &  43.9±1.6 \\\cdashline{3-8}
 & \multirow{2}{*}{\textcolor{cyan}{shuffle words  }   }   &  20.1±2.7 &   4.20±2.5 &  18.1±2.0 &   7.80±1.6 &  14.2±2.0 &   8.60±0.5 \\
\rowfont{\color{gray}} & & 59.4±1.8 &  46.5±0.9 &  46.7±0.8 &  49.7±2.2 &  50.5±2.0 &  41.5±0.4  \\\cdashline{3-8}
 & \multirow{2}{*}{\textcolor{olive}{repeat sentences } }   &  51.1±0.8 &  46.4±1.0 &  34.8±1.0 &  37.6±4.0 &  16.4±2.3 &  41.3±1.4 \\
\rowfont{\color{gray}} & & 61.4±0.5 &  48.0±0.9 &  48.4±0.9 &  53.7±3.3 &  51.1±1.2 &  48.5±1.1  \\\cdashline{3-8}
 & \multirow{2}{*}{\textcolor{olive}{shuffle sentences}}    &  49.9±1.2 &  46.6±1.3 &  34.0±1.0 &  42.4±0.1 &  16.4±1.5 &  40.0±0.9 \\
\rowfont{\color{gray}} & & 62.2±1.2 &  49.6±0.8 &  48.6±1.4 &  54.4±0.2 &  50.1±0.7 &  45.2±1.7 \\\cdashline{3-8}
 & \multirow{2}{*}{\textcolor{blue}{shuffle instructions}} &   2.40±2.6 &   3.00±3.9 &   4.20±2.2 &   2.70±3.7 &   1.80±1.3 &   1.20±1.5 \\
\rowfont{\color{gray}} & & 53.3±1.4 &  45.9±1.1 &  41.1±1.1 &  33.8±1.1 &  38.2±1.0 &  28.3±1.0\\
\hline
{} & &       DTT &       GEC &        TG &        KT &        OE &        QR \\
\hline

\multirow{18}{*}{\rotatebox{90}{\begin{tabular}{c} \rougel\end{tabular}}}  & \multirow{2}{*}{original instru.   }         &  31.6±0.4 &  83.1±1.4 &  33.8±0.6 &   56.3±1.7 &  24.4±2.8 &  48.2±0.4 \\
\rowfont{\color{gray}}& & 43.3±0.5 &  85.7±1.1 &  37.4±0.5 &  64.8±1.9 &  32.7±1.4 &  68.0±0.2  \\\cdashline{3-8}
 & \multirow{2}{*}{\textcolor{cyan}{delete stopwords }  }  &  27.8±0.4 &  83.3±0.6 &  31.5±0.3 &   49.9±1.1 &  22.6±0.6 &  37.2±0.2 \\
\rowfont{\color{gray}} & & 43.8±0.4 &  85.2±0.5 &  37.2±0.3 &  63.6±0.3 &  34.2±1.0 &  68.0±0.3  \\\cdashline{3-8}
 & \multirow{2}{*}{\textcolor{cyan}{delete [5,10] words  }}    &  29.9±1.3 &  78.2±8.8 &  31.0±0.8 &   49.7±4.6 &  24.0±0.8 &  41.0±2.3 \\
\rowfont{\color{gray}} & & 42.2±0.5 &  85.0±2.0 &  36.8±0.2 &  64.0±1.0 &  34.4±1.8 &  67.5±0.0  \\\cdashline{3-8}
 & \multirow{2}{*}{\textcolor{cyan}{insert [5,10] words  }}    &  30.9±0.3 &  81.2±1.2 &  33.2±0.7 &   54.2±1.1 &  25.3±0.5 &  46.8±0.4 \\
\rowfont{\color{gray}} & & 42.3±0.5 &  86.6±0.4 &  37.0±0.2 &  65.8±1.2 &  36.7±1.9 &  67.3±0.3  \\\cdashline{3-8}
 & \multirow{2}{*}{\textcolor{cyan}{replace [5,10] words   } } &  31.3±0.9 &  82.6±0.7 &  30.9±0.7 &   53.5±2.8 &  26.9±1.2 &  42.9±3.6 \\
 \rowfont{\color{gray}}& & 42.5±0.4 &  84.0±1.5 &  36.9±0.4 &  65.6±1.6 &  36.0±3.3 &  68.2±0.5 \\\cdashline{3-8}
 & \multirow{2}{*}{\textcolor{cyan}{shuffle words     } }  &  20.7±2.6 &  71.1±5.2 &  21.9±0.5 &   33.9±2.0 &  22.9±1.0 &  24.3±1.3 \\
 \rowfont{\color{gray}}& & 44.9±0.4 &  84.6±0.7 &  36.3±0.4 &  62.4±0.2 &  33.8±3.6 &  67.7±0.1 \\\cdashline{3-8}
 & \multirow{2}{*}{\textcolor{olive}{repeat sentences   } } &  33.1±0.5 &  83.5±0.2 &  31.6±0.6 &   52.0±3.1 &  26.9±0.6 &  52.0±0.4 \\
 \rowfont{\color{gray}}& & 42.9±0.4 &  85.9±0.3 &  37.4±0.3 &  64.8±1.6 &  36.8±2.5 &  67.3±0.3  \\\cdashline{3-8}
 & \multirow{2}{*}{\textcolor{olive}{shuffle sentences }  } &  31.1±0.2 &  83.6±2.3 &  33.9±0.5 &   55.1±1.2 &  25.8±0.6 &  44.6±2.3 \\
 \rowfont{\color{gray}}& & 41.9±0.3 &  86.4±0.7 &  37.1±0.9 &  64.6±1.2 &  38.2±1.8 &  67.6±0.4  \\\cdashline{3-8}
 & \multirow{2}{*}{\textcolor{blue}{shuffle instructions}} &   9.90±2.2 &   7.20±8.4 &  10.1±1.4 &  18.8±11.3 &  15.0±6.0 &  16.7±5.8 \\
 \rowfont{\color{gray}}& & 42.0±0.3 &  87.2±0.6 &  27.5±0.4 &  52.2±0.4 &  31.4±3.0 &  66.0±0.4  \\
\hline
\end{tabu}
\caption{The \shortinstruc~(in black) and \longinstruc~(in gray) results of the T5-3B model on the \emph{119 English} $test$ tasks of \naturaltwo~by different levels of perturbations. We use different colors to denote perturbations in \textcolor{cyan}{word level}, \textcolor{olive}{sentence level} and \textcolor{blue}{instruction level}.}
\label{tab:english_3B}
\end{table*}

\begin{table*}[!ht]
 \setlength{\abovecaptionskip}{5pt}
\centering

\begin{tabu}{l|l|lllllll}
\hline
{metric} & &        AC &       CEC &        CR &       DAR &        WA &        TE \\
\hline
\multirow{9}{*}{\rotatebox{90}{\begin{tabu}{c} \exactmatch\end{tabu}}} & \multirow{1}{*}{original instru.   }          &  68.00 &  54.00 &  57.86 &  63.14 &  47.50 &  64.54 \\
 & \multirow{1}{*}{\textcolor{cyan}{delete stopwords} }    &  54.15 &  53.14 &  54.93 &  52.86 &  43.00 &  61.79 \\
 & \multirow{1}{*}{\textcolor{cyan}{delete [5,10] words } }    &  65.46 &  54.00 &  54.29 &  57.29 &  46.50 &  61.33 \\
 & \multirow{1}{*}{\textcolor{cyan}{insert [5,10] words}  }    &  59.62 &  54.29 &  57.29 &  56.00 &  47.50 &  64.12 \\
 & \multirow{1}{*}{\textcolor{cyan}{replace [5,10] words } }   &  64.77 &  54.43 &  55.57 &  56.71 &  47.75 &  58.67 \\
 & \multirow{1}{*}{\textcolor{cyan}{shuffle words  }   }   &   47.23 &  53.86 &  51.07 &  39.86 &  41.88 &  56.12 \\
 & \multirow{1}{*}{\textcolor{olive}{repeat sentences } }   &  67.08 &  53.86 &  57.64 &  64.71 &  48.88 &  64.17 \\
 & \multirow{1}{*}{\textcolor{olive}{shuffle sentences}}    &  66.69 &  54.14 &  57.43 &  66.14 &  47.25 &  64.08 \\
 & \multirow{1}{*}{\textcolor{blue}{shuffle instructions}} &   43.69 &  44.14 &  49.64 &  38.00 &  39.50 &  48.29 \\
\hline
{} & &       DTT &       GEC &        TG &        KT &        OE &        QR \\
\hline

\multirow{9}{*}{\rotatebox{90}{\begin{tabular}{c} \rougel\end{tabular}}}  & \multirow{1}{*}{original instru.   }         &  49.17 &  87.80 &  39.11 &  67.78 &  60.35 &  66.25 \\
 & \multirow{1}{*}{\textcolor{cyan}{delete stopwords }  }  &  49.56 &  87.19 &  38.21 &  65.07 &  54.82 &  66.46 \\
 & \multirow{1}{*}{\textcolor{cyan}{delete [5,10] words  }}    &  49.26 &  87.69 &  38.15 &  65.94 &  58.24 &  66.19 \\
 & \multirow{1}{*}{\textcolor{cyan}{insert [5,10] words  }}    &  49.28 &  87.55 &  38.62 &  66.17 &  57.08 &  65.98 \\
 & \multirow{1}{*}{\textcolor{cyan}{replace [5,10] words   } } &  49.91 &  88.77 &  38.32 &  66.46 &  57.68 &  65.42 \\
 & \multirow{1}{*}{\textcolor{cyan}{shuffle words     } }  &  49.39 &  87.33 &  36.90 &  65.89 &  52.85 &  67.77 \\
 & \multirow{1}{*}{\textcolor{olive}{repeat sentences   } } &  49.26 &  88.81 &  38.76 &  67.34 &  60.19 &  66.33 \\
 & \multirow{1}{*}{\textcolor{olive}{shuffle sentences }  } &  49.41 &  87.80 &  39.05 &  66.86 &  59.17 &  66.17 \\
 & \multirow{1}{*}{\textcolor{blue}{shuffle instructions}} &   48.93 &  86.32 &  31.08 &  62.11 &  38.08 &  64.82 \\
\hline
\end{tabu}
\caption{The \longinstruc~results of the T5-11B model on the \emph{119 English} $test$ tasks of \naturaltwo~by different levels of perturbations. We use different colors to denote perturbations in \textcolor{cyan}{word level}, \textcolor{olive}{sentence level} and \textcolor{blue}{instruction level}.}
\label{tab:english_11B}
\end{table*}

\section{Experiment}
This work studies the robustness of learning from task instructions from three scenarios: (i) \textit{the system robustness when dealing with instructions that are deliberately perturbed} (Section \ref{sec:perturbed}); (ii) \textit{the system robustness when a task confronts diverse instructions that are written by different annotators} (Section \ref{sec:rephrase}); (iii) \textit{the system robustness when it tests on tasks that have instructions with distinct abstractiveness} (Section \ref{sec:complexity}). The following subsections present our experiments about the three dimensions of robustness study, respectively.

\paragraph{The pretrained instruction learning model.} To test the robustness of a pretrained model that learned NLP tasks from instructions, we choose the T5 model \cite{DBLPRaffelSRLNMZLL20} with 3B and 11B parameters finetuned on the training set of \naturaltwo~\cite{DBLP07705}. 

\naturaltwo~
consists of 1,616 NLP tasks in total and each task is explained by four items: \texttt{Definition} (a short paragraph describing the task semantics), \texttt{Positive Examples} (a couple of examples with inputs, correct outputs, and explanations), \texttt{Negative Examples} (a couple of examples with inputs, incorrect outputs, and explanations), and \texttt{Instances} (a large set of  instances with inputs and correct outputs). One sample is shown in Table \ref{tab:naturalv2}. 
As far as we know, \naturaltwo~is the largest dataset containing \texttt{human-oriented} task instructions, and T5 denotes one of the state-of-the-art PLM models (we do not consider  GPT-3 due to the hardware constraints). As the data for training the instruction learning model, the $train$ of \naturaltwo~has 1,462 tasks.

\subsection{Robustness when instructions are perturbed}\label{sec:perturbed}

\paragraph{Dataset.} 

This experiment uses the $test$ of \naturaltwo, which has  154 tasks (119 English tasks and 35 tasks in other languages).  The robustness of the pretrained T5 model on $train$ is tested on this $test$ where instructions are perturbed. All $test$ tasks are clustered into 12 categories with the official evaluation metrics either \exactmatch~or \rougel~\cite{Lin2004-wx}. Table \ref{tab:abbreviation} lists all the 12 categories and their respect evaluation metrics.

Note that \citet{DBLP07705} concatenated \texttt{Task name}, \texttt{Definition}, \texttt{Positive Examples} and \texttt{Negative Examples} in various combinations as a long instruction, referred to as ``Instruction with Examples'' (\longinstruc). This paper uses the definition and two positive examples as \longinstruc~ configuration. In order to study the robustness when those examples are available or not, we report \longinstruc~as well as \shortinstruc~which uses only the \texttt{Definition} as instructions.

\begin{table*}[ht]
\centering
\begin{tabu}{l|p{11cm}}
\hline
\multirow{1}{*}{Original Instruction} & In this task, you will be shown an incorrect English sentence. You need to generate a corrected form of the input sentence.\\
\hline
\multirow{1}{*}{New Instruction 1} & Given an incorrect English sentence. Your task is to correct the input sentence and write out his correct form.\\
\hline
\multirow{1}{*}{New Instruction 2} & You will see an erroneous English sentence in this task. You must create the input sentence in its corrected form.\\
\hline
\multirow{1}{*}{New Instruction 3} & You are given an English sentence that has some small grammatical errors. Your task is to output the given sentence and fix all the errors inside. The output should be a grammatically correct sentence.\\
\hline
\hline
\end{tabu}

\caption{An example in our constructed dataset where four diverse instructions are prepared for every single task.}
\label{tab:four_instructions}
\end{table*}

\begin{table*}[ht]
\centering

\begin{tabu}{l|l|l|llllll}
\hline
\multirow{6}{*}{\rotatebox{90}{\begin{tabu}{c} T5-3B\end{tabu}}} & &    &    AC &       CEC &        CR &       DAR &        WA &        TE \\
\cline{2-9}
& \multirow{2}{*}{EM} & orig. &  69.5  &  69.2  &  58.5  &  48.7  &  44.5  &  44.8  \\
&  & reph.      &  67.3±1.0 &  69.3±0.2 &  56.8±1.6 &  43.3±2.7 &  43.2±1.9 &  52.7±5.8 \\\cline{2-9}
&  & &       DTT &       GEC &        TG &        KT &        OE &        QR \\
\cline{2-9}
& \multirow{2}{*}{\rougel}  & orig. &  44.7  &  85.0  &  25.5  &  73.2  &  34.7  &  81.0    \\
 & & reph.      &  44.2±0.3 &  84.3±0.3 &  24.6±0.2 &  72.0±0.6 &  32.7±0.1 &  81.3±0.2 \\
\hline

\multirow{6}{*}{\rotatebox{90}{\begin{tabu}{c} T5-11B\end{tabu}}} & &    &    AC &       CEC &        CR &       DAR &        WA &        TE \\
\cline{2-9}
& \multirow{2}{*}{EM} & orig. &  63.5 &  74.0 &   62.0 &  51.5 &   51.5 &  67.0 \\
&  & reph.      &  61.8±0.6 &  74.5±1.1 &  56.7±5.3 &  50.3±4.1 &  50.0±0.4 &  65.0±2.4  \\
\cline{2-9}
&  & &       DTT &       GEC &        TG &        KT &        OE &        QR \\
\cline{2-9}
& \multirow{2}{*}{\rougel}  & orig. &  48.8 &  85.0 &  26.4 &  71.5 &  52.1 &  75.0   \\
&  & reph.      &  48.7±0.5 &  85.4±0.2 &  26.3±0.2 &  71.6±0.5 &  51.6±1.0 &  76.9±2.1 \\
\hline

\end{tabu}

\caption{The results of the T5-3B\&11B models on our \texttt{Para-Instructions} dataset.}
\label{tab:english_new_instruction_3B}
\end{table*}

\paragraph{Instruction perturbation.}
To imitate how instructions are perturbed in the real world, we design various perturbation methods that try to change the original instruction by its surface form or even semantics. Table \ref{tab:perturbation_examples} summarizes all the  perturbation  methods this work uses, including \textbf{word-level perturbations}, such as i) delete stop words; ii) randomly delete five to ten words; iii) insert masks at random positions and replace with words predicted by pretrained BERT \cite{Devlin2018-cl}; iv) randomly replace words by the predictions of pretrained BERT, and v) shuffle the words in the instruction, and \textbf{sentence-level perturbations}, such as i) repeating a random sentence, and ii) shuffle the sentences in the instruction, and \textbf{instruction-level perturbation}, where we try to replace the original task definition with the definition of a randomly chosen task.

Spacy \cite{spacy} is used to remove stop words from instructions and to split instructions into separate sentences, and BERT is used to insert and replace words in instructions.

\paragraph{Results.}
Each metric was run three times, and the mean and standard deviation of the three runs were reported. 
Tables \ref{tab:english_3B}\&\ref{tab:xlingual_3B} presents the performance on English and non-English subsets of \naturaltwo's $test$, respectively. Each table summarizes the perturbation results by word-level, sentence-level, and instruction-level for both the \shortinstruc~and \longinstruc~setups. Our goal is to study i) the effects of different levels of perturbations; ii) the effects of including positive examples in the instructions; iii) whether there are any distinct phenomena between English tasks and non-English tasks or not. From the two tables, we observe that:

 \textbullet\enspace \textbf{Effects of different levels of perturbations in \shortinstruc}: i) word-level perturbations  degrade the performance slightly, among which the ``shuffle words'' has the worst effect; ii) generally, sentence-level perturbations have less influence than the word-level perturbations: we only notice tiny drop compared with ``original instructions'' and no clear difference between ``repeat sentences'' and ``shuffle sentences'' is observed; iii) ``shuffle instructions'' doesn't work; this is within expectation since a random instruction  provides no useful supervision for the target task.

    \textbullet\enspace 
    \textbf{When the instructions contain examples (i.e., \longinstruc)}: i) the system is pretty robust among all perturbations; ii) ``shuffle instructions'' (note that the examples are still specific to the target task) only decreases the performance to a small extent, which indicates that \textit{``correct examples+incorrect instruction'' is better than ``correct instruction without examples''}. In other words, \textcolor{chocolate}{providing a couple of examples in the instruction rather than merely a short piece of task description is still the key to instruction-driven learning}.

    \textbullet\enspace \textbf{Does larger models lead to higher robustness?} We report \longinstruc~results of T5-11B in Table \ref{tab:english_11B}. We only observe  occasional performance drops by word-level perturbations in a few task categories, such as CR, DAR, and OR. In general, only ``shuffle instruction'' can lead to clear deterioration while other perturbations do not have a noticeable effect.
     
    \textbullet\enspace \textbf{Difference between English and non-English}: To save space, we move the results of non-English $test$ tasks to the Table \ref{tab:xlingual_3B} in Appendix \ref{app:3BonXling}. By comparing Table \ref{tab:english_3B} and Table \ref{tab:xlingual_3B}, we do not notice any different phenomena in non-English tasks.

\paragraph{How to interpret the phenomena caused by instruction perturbations?} Based on the aforementioned observations, we conclude that perturbations,  retaining as much task semantic as the original instructions, can yield less influence on the system's robustness. In other words, two instructions that are more likely to be paraphrased tend to have similar performance. We further validate this argument in  the subsequent Section \ref{sec:rephrase}.

Since \longinstruc~generally performs better than \shortinstruc, hereafter, we report \longinstruc~for the remaining experiments.

\begin{table}[ht]
\centering
\begin{tabu}{l|l|l|l}
\hline
 & Task & T5-3B & T5-11B\\
\hline
\multirow{14}{*}{\rotatebox{90}{\begin{tabu}{c} \exactmatch\end{tabu}}} 
&sum                     &     0.00±0.0   &  11.4±4.2\\
&synonyms                &     1.00±0.0   &  2.40±0.9 \\
&rhymes                  &     1.00±0.0   &  0.00±0.0\\
&larger animal           &     5.00±2.5   &  45.8±7.6\\
&diff                    &     1.10±0.9   &  3.30±1.1\\
&first word letter       &     1.00±0.0   &  46.0±9.3\\
&second word letter      &     1.30±0.5   &  19.9±6.5\\
&singular to plural      &     0.80±0.4   &  55.0±1.2\\
&antonyms                &     13.6±1.6   &  33.1±0.6\\
&word in context         &     9.00±12.2  &  31.7±0.0\\
&letters list            &     2.00±0.0   &  1.80±0.6\\
&sentence similarity     &     0.00±0.0   &  0.00±0.0\\
&sentiment               &     34.1±8.9   &  49.0±2.8\\
&num to verbal           &     0.00±0.0   &  0.00±0.0\\
&orthography  &     1.20±0.4   &  7.00±2.3\\
\hline
\multirow{10}{*}{\rotatebox{90}{\begin{tabular}{c} \rougel\end{tabular}}}
&translation en-es       &     8.50±0.5   &  23.3±2.8\\
&translation en-de       &     9.50±1.0   &  10.1±5.8\\
&translation en-fr       &     12.4±0.8   &  37.0±0.8\\
&informal to formal      &     30.3±0.4   &  32.8±2.3\\
&negation                &     70.7±11.6  &  71.6±6.1\\
&cause and effect        &     50.5±15.1  &  31.8±9.7\\
&taxonomy animal         &     24.0±3.9   &  26.6±7.0\\
&common concept          &     0.00±0.0   &  1.70±2.2\\
&active to passive       &     60.3±2.4   &  94.4±3.5\\

\hline
\end{tabu}
\caption{Train T5-3B\&11B on \naturaltwo~and test it on \diverseprompt~ \cite{Honovich_undated-gv}. }
\label{tab:crossdata_3B}
\end{table}

\begin{table*}[!ht]
    \centering
\begin{tabu}{l|c|p{11cm}}
\hline
       &  \rougel &    prompt\\
\hline
\multirow{6}{*}{\rotatebox{90}{\begin{tabular}{c} negation\end{tabular}}}  & \multirow{1}{*}{63.32} &                      For each input, write a sentence that expresses the exact opposite meaning of the input. \\\cdashline{3-3}
& 65.95 &                                                    Change the fact stated in the sentence to an opposite fact. \\\cdashline{3-3}
& \multirow{1}{*}{66.50}  & You will be given a sentence that states a fact (that might be true or not). Try to state the opposite fact. \\\cdashline{3-3}
& 68.92  &                                                                               Negate the following sentence: \\\cdashline{3-3}
& 74.05  &                                                                             output the negation of the input \\\cdashline{3-3}
& 75.39  &                                                                                    Negate the given sentence \\\cdashline{3-3}
& 75.87  &                                                                                           write the negation \\\cdashline{3-3}
& 82.83  &                                                                Write a negated version of the given sentence \\\hline

\multirow{12}{*}{\rotatebox{90}{\begin{tabular}{c} cause \& effect\end{tabular}}} & 16.26  &                                                                               Which of the two events is the cause? \\\cdashline{3-3}
& 21.23  &                                                                      Which of the following sentences is the cause? \\\cdashline{3-3}
& 21.97  &                                               Output the cause (other sentence describes what happened as a result) \\\cdashline{3-3}
& \multirow{1}{*}{28.71}  &                         Output the sentence describing the cause (the other sentence is what happened as a result). \\\cdashline{3-3}
& \multirow{1}{*}{32.17}  & Each input consists of two sentences, where one is the cause and the other is the outcome. Write the cause sentence \\\cdashline{3-3}
& \multirow{1}{*}{39.03}  &                        The input consists of two sentences. One is the cause of the other. Write the cause sentence \\\cdashline{3-3}
& 40.30  &                                                               Find the cause in the following cause and effect pair \\\cdashline{3-3}
& 41.83  &                                                                   The input is a cause and effect. Write the cause. \\\cdashline{3-3}
& 44.78 &                                                                    The input is a cause and effect, write the cause \\\hline

\multirow{9}{*}{\rotatebox{90}{\begin{tabular}{c} taxonomy animal\end{tabular}}} & 12.06 &                                   extract animals \\\cdashline{3-3}
&21.46 &          List which of the following are animals \\\cdashline{3-3}
&22.62 &                     Find the animals in the list \\\cdashline{3-3}
&24.08 &             List the animals from the given words \\\cdashline{3-3}
&27.86 &   find the animals in the following list of words \\\cdashline{3-3}
&28.25 &          write all animals from the list of words \\\cdashline{3-3}
&33.88 &     write only the animals from the list of words \\\cdashline{3-3}
&33.89 &                 Extract all animals from the list \\\cdashline{3-3}
&35.32 &           Extract all animals from the input list \\\hline

\multirow{8}{*}{\rotatebox{90}{\begin{tabular}{c} active\_to\_passive \end{tabular}}} & 88.18 &                       Write the following sentence in passive language \\\cdashline{3-3}
&88.63 &                                   output the passive form of the input \\\cdashline{3-3}
&95.71 &                For each input, write the passive form of the sentence. \\\cdashline{3-3}
&95.87 &                                   Rewrite the sentence in passive form \\\cdashline{3-3}
&96.08 &                            Change the sentence from active to passive. \\\cdashline{3-3}
&96.21 &                             Rewrite the input sentence in passive form \\\cdashline{3-3}
&96.92 &   Change the wording of the following sentence from active to passive. \\\cdashline{3-3}
&97.80 &                  turn the sentence from active    tense to passive tense. \\\hline

\end{tabu}
\caption{\rougel~per prompt for four tasks in \diverseprompt.}
\label{tab:perperprompt}
\end{table*}

\subsection{Robustness when instructions are rephrased}\label{sec:rephrase}
In addition to the instruction perturbation experiments which may lead to grammatically incorrect instructions, this subsection studies a scenario where multiple valid instructions are available for the same task. This is a real-world challenge since the end users are free to write any open-form instructions to guide the system. 

\paragraph{Data construction.} Unfortunately, there are no existing datasets that contain tasks with diverse \texttt{human-oriented} instructions.  To build a dataset,  we randomly choose 2 $English$\footnote{We considered English only since it is hard to find annotators who are familiar with many non-English languages.} tasks from each of the 12 categories in the $test$ of \naturaltwo; this results in 23 tasks\footnote{The ``grammar error 
correction''  has only one task.} that are uniformly distributed over those categories.  We hire three  graduate students in the NLP area to construct a dataset of diverse instructions with the following two steps:

\textbullet\enspace \textbf{Writing step}: for each task, each annotator is provided with 10 labeled instances, the task name, and the original task \texttt{Definition}. Then each annotator is required to write a new definition based on her/his understanding of the task. After this step, each task will have 3 new versions of raw instructions.

\textbullet\enspace \textbf{Validation step}: For each raw instruction, we forward it to two  checkers who judge if it is valid for the task and if it is different enough from the original \texttt{Definition}. If any issues are found, the original author will be asked to modify the instruction. This process is repeated until both checkers are satisfied with the new instruction. After this step, each task will have 4 versions of  high-quality instructions (one original, three new).

Table \ref{tab:four_instructions} lists  an example task  with 4 instructions in our constructed data. Compared with the perturbed instructions demonstrated in Table \ref{tab:perturbation_examples}, it is clear that the tasks in our new dataset are equipped with more valid instructions and they are more like paraphrases of each other while conveyed with more distinct textual expressions.
Our constructed dataset, ``\texttt{Para-Instructions}'', as the first one that contains multiple \texttt{human-oriented} instructions for each task, is released to the community.

\paragraph{Results.} For all three new instructions, we report their averages and standard deviations. Table \ref{tab:english_new_instruction_3B} compares  the performance of the original instruction and the three new instructions by T5-3B and T5-11B models. We notice that the system is considerably robust among all four versions of instructions. \textcolor{chocolate}{It suggests that providing a paraphrasing instruction can generally keep the performance once demonstrations are available}.

\subsection{Robustness when instructions have different abstractiveness}\label{sec:complexity}

Another challenge in instruction-driven task generalization is to encourage the same system to work well on datasets that contain instructions constructed from totally different templates or abstractiveness. This phenomenon often exists between two different datasets. For example, \naturaltwo~has instructions that are mostly a paragraph while \diverseprompt~\cite{Honovich_undated-gv} contains prompts only---mostly short and abstractive sentences.

\paragraph{Dataset.} To check the robustness of the  model finetuned on \naturaltwo, we test it on the \diverseprompt~dataset. \diverseprompt~has 24 tasks and each task has an average of 7.75 prompts. \diverseprompt~was the only dataset that provides multiple prompts for a task before we wrote this paper, and the prompts are much more concise (average word size 10.75) than the training instructions of \naturaltwo. 

\paragraph{Results.} Table \ref{tab:crossdata_3B} presents the results on \diverseprompt~by the model finetuned on \naturaltwo. For this experiment, we observe a totally different story: 

\textbullet\enspace First, the pretrained model on \naturaltwo~ does not work well on \diverseprompt. Regardless of the task types (shown by the evaluation metrics: \exactmatch~or \rougel), most tasks show unsatisfactory performance. This can result from two reasons: on the one hand, the instruction format (i.e., prompts in \diverseprompt) is significantly different from that in \naturaltwo---\diverseprompt~has prompts that mostly contain a few words while instructions in \naturaltwo~are more specific, informative and are human-oriented; on the other hand, the tasks in \diverseprompt~are very distinct from those in \naturaltwo, such as ``sum'', ``diff'', ``first word letter'', ``common concept'', etc.

\textbullet\enspace Second, even for some tasks that co-exist in \naturaltwo~and \diverseprompt, e.g., ``word in context'', ``translation'', ``sentiment'', etc., their performances  fall short of expectations. For example, ``Text Generation'' (TG) in Table \ref{tab:english_new_instruction_3B} gets consistent performance around 24.6 by \rougel, whereas the translation tasks in \diverseprompt~ranging from 8 to 13. 

\textcolor{chocolate}{The preceding two items hint at the fragility of the model when dealing with unseen tasks and unfamiliar instruction formats, despite its training on more than 1,000 instruction-laden tasks. So far, we still lack a method to effectively learn instruction to solve tasks in the absence of demonstration.} The majority of instruction-tuning datasets commonly furnish only a single instruction for each input-output pair, aiming to enhance the system's capacity for instruction following through an increased number of training tasks. However, this approach may prove suboptimal in supervising a system that can effectively generalize to unfamiliar tasks with unfamiliar formats of instructions.

\textbullet\enspace Third, it is worth mentioning that this \diverseprompt~dataset does not provide demonstrations. In this case, we notice mostly \textit{greater standard deviations} in Table \ref{tab:crossdata_3B} than that in Table \ref{tab:english_new_instruction_3B}. \textcolor{chocolate}{This indicates that even paraphrasing prompts can not guarantee consistent performance if no demonstrations exist.} Then we dive into the performance per prompt for tasks that have larger standard deviations of \rougel, such as ``negation'', ``cause and effect'', ``taxonomy animal'' and ``active to passive'', in Table \ref{tab:perperprompt}. We found the following factors that affect prompt performance in text generation tasks: \textcolor{chocolate}{(i) it is better to use a verb that clearly has a single sense. For example, ``write'' and ``extract'' are generally better than ``output'', ``list'', which can be verbs or nouns; (ii) explicitly presenting the action is better than that implicitly. For example, ``$\cdots$, write the $\cdots$'' is better than a question ``which of the $\cdots$ is $\cdots$?''.}

This story shows that there is still a long way to go for instruction-driven task generalization, especially when the system deals with less familiar tasks with abstractiveness-distinct instructions.

\section{Conclusion}
This paper proposed a new issue, the robustness of instruction-tuned models. Experiments show that the instruction-tuned model may not learn the content of the instructions itself in some tasks, but rather the pattern of positive examples. The robustness of the instruction-tuned model is very important, which determines the stability and reliability of instruction fine-tuning models in real-world applications. Using only instructions to tell models what they should do is a long-term goal of AI, and we hope that future work will be done to improve the robustness of the instruction-tuned model.

\section*{Acknowledgments}

The authors would like to thank all the anonymous reviewers for their insightful comments and suggestions. 

\section*{Limitations}
We summarize the limitations of this work as follows:
\begin{itemize}
    \item First, it is still challenging to figure out why a particular instruction results in better or worse performance. We tried to explain a part of the observations, but some of the phenomena are still mysterious.
    \item Technically we can create a much larger \texttt{Para-Instruction} dataset than 23 that provides diverse instructions for NLP tasks. Due to the budget limitation, we only collected 23 tasks.
    \item Some prior work showed that larger PLMs often bring better performance in dealing with instruction learning. Due to hardware constraints, we are unable to try PLMs that are larger than T5-11B.
\end{itemize}

\section*{Ethics Statement}
We do not anticipate any ethical issues particularly to the topics of this work. Nevertheless, some work presented here  extensively uses large-scale pretrained models with self-attention, which may lead to substantial financial and environmental costs.


\bibliography{anthology,custom}
\bibliographystyle{acl_natbib}

\appendix
\begin{table*}[ht!]
\centering
\begin{tabu}{l|lll|l}
\hline
& \multicolumn{3}{c|}{\exactmatch} & \rougel\\
&       AC &       CEC &        TE     &  TG\\
\hline
\multirow{2}{*}{original   }          &   76.3±3.4 &  43.1±0.7 &  19.7±1.4  &10.3±2.0 \\
\rowfont{\color{gray}}& 72.0±0.8 &  54.2±0.6 &  20.0±1.2  & 8.80±0.7 \\
 \multirow{2}{*}{\textcolor{cyan}{delete stopwords} }    &   69.0±4.3 &  20.8±0.6 &  10.1±0.2    &5.70±2.2 \\
 \rowfont{\color{gray}}& 69.7±1.9 &  54.8±0.2 &  19.0±1.4 &  6.30±0.8 \\
 \multirow{2}{*}{\textcolor{cyan}{delete [5,10] words}  }    &  56.3±2.8 &  28.9±2.2 &  11.2±4.9    &9.70±1.6 \\
 \rowfont{\color{gray}}& 73.0±2.9 &  54.6±0.6 &  21.0±1.9   & 8.40±0.1 \\
 \multirow{2}{*}{\textcolor{cyan}{insert [5,10] words }  }   &   72.0±2.2 &  35.5±1.4 &  17.6±3.4   &9.50±0.6 \\
 \rowfont{\color{gray}}& 69.3±4.1 &  55.7±0.3 &  19.6±0.6   & 7.20±1.0 \\
 \multirow{2}{*}{\textcolor{cyan}{replace [5,10] words} }    &   68.3±0.5 &  39.7±1.5 &  17.8±3.1   &10.5±1.6 \\
 \rowfont{\color{gray}}& 67.0±7.9 &  54.2±0.3 &  17.4±0.8   & 6.80±1.9 \\
\multirow{2}{*}{\textcolor{cyan}{shuffle words  }  }    &    3.00±1.6 &   0.20±0.1 &   1.40±2.0    &8.60±2.3 \\
\rowfont{\color{gray}}& 70.0±5.7 &  53.2±0.1 &  16.0±2.6   & 8.30±1.7 \\
 \multirow{2}{*}{\textcolor{olive}{repeat sentences} }    &   73.3±2.6 &  32.6±2.7 &  14.6±3.5    &9.20±1.7 \\
 \rowfont{\color{gray}}& 74.7±1.2 &  55.2±0.6 &  20.9±2.4   & 9.00±0.7 \\
 \multirow{2}{*}{\textcolor{olive}{shuffle sentences} }   &   69.7±3.1 &  34.8±1.7 &  17.3±0.9   &5.80±1.7 \\
 \rowfont{\color{gray}}& 67.3±1.2 &  54.9±0.5 &  19.1±1.6   & 7.00±1.8 \\
 \multirow{2}{*}{\textcolor{blue}{shuffle instructions} }&    0.00±0.0 &  14.9±1.7 &   0.00±0.0    &0.60±0.9 \\
 \rowfont{\color{gray}}& 40.0±2.9 &  52.6±0.4 &   5.30±0.7   & 1.90±1.5 \\
\hline
\end{tabu}
\caption{The \shortinstruc~(in black) and \longinstruc~(in gray) results of the T5-3B model on the \emph{35 non-English} $test$ tasks of \naturaltwo~by different levels of perturbations. }

\label{tab:xlingual_3B}
\end{table*}
\begin{table*}[ht!]
\centering
\begin{tabu}{l|lll|l}
\hline
& \multicolumn{3}{c|}{\exactmatch} & \rougel\\
&       AC &       CEC &        TE     &  TG\\
\hline
\multirow{1}{*}{original   }          &   86.00 &  69.57 &  32.00  &13.27 \\
 \multirow{1}{*}{\textcolor{cyan}{delete stopwords} }    &   78.0 &  69.17 &  28.00 & 14.00\\
 \multirow{1}{*}{\textcolor{cyan}{delete [5,10] words}  }    &  85.0 &  69.07 &  28.00 &  13.00 \\
 \multirow{1}{*}{\textcolor{cyan}{insert [5,10] words }  }   &   87.0 &  69.27 &  30.00 &  12.83 \\
 \multirow{1}{*}{\textcolor{cyan}{replace [5,10] words} }    &    85.0 &  67.47 &  30.67 &  13.83 \\
\multirow{1}{*}{\textcolor{cyan}{shuffle words  }  }    &   76.0 &  68.03 &  29.67 &  13.17 \\  
 \multirow{1}{*}{\textcolor{olive}{repeat sentences} }    &   87.0 &  69.53 &  31.00 &  13.27 \\
 \multirow{1}{*}{\textcolor{olive}{shuffle sentences} }   &  85.0 &  69.97 &  29.33 &  13.27 \\
 \multirow{1}{*}{\textcolor{blue}{shuffle instructions} }&    60.0 &  69.07 &  19.67 &   4.47 \\
\hline
\end{tabu}
\caption{The \longinstruc~results of the T5-11B model on the \emph{35 non-English} $test$ tasks of \naturaltwo~by different levels of perturbations. }

\label{tab:xlingual_11B}
\end{table*}

\section{Finetuning Details}
In this work, we finetune T5 using the HuggingFace library. We use a scheduler with a constant learning rate of 5e-05. All models are optimized by AdamW~\cite{AdamW} with a batch size of 8. T5-3B is fine-tuned on one NVIDIA RTX A5000 GPU and T5-11B on four NVIDIA A100 GPUs. DeepSpeed\footnote{https://github.com/microsoft/DeepSpeed} is used to parallelize the models, and bfloat16 precision is enabled to save GPU memory.

\section{T5-3B Results on Non-English Tasks}
\label{app:3BonXling}

Table \ref{tab:xlingual_3B} presents the \shortinstruc~(in black) and \longinstruc~(in gray) results of the T5-3B model on the \emph{35 non-English} $test$ tasks of \naturaltwo~by different levels of perturbations.

\section{T5-11B Results on Non-English Tasks}
\label{app:11BonXling}

Table \ref{tab:xlingual_11B} presents the \longinstruc~results of the T5-11B model on the \emph{35 non-English} $test$ tasks of \naturaltwo~by different levels of perturbations.



\end{document}